# Remote Sensing ChatGPT: Solving Remote Sensing Tasks with ChatGPT and Visual Models

*Haonan Guo[1], Xin Su*[2], Chen Wu[1], Bo Du [3], Liangpei Zhang[1], Deren Li[1]*
1 State Key Laboratory of Information Engineering in Surveying, Mapping and Remote Sensing, Wuhan University, Wuhan, China
2 School of Remote Sensing and Information Engineering, Wuhan University, Wuhan, China
3 School of Computer Science, Wuhan University, Wuhan, China

**ABSTRACT**

Recently, the flourishing large language models(LLM), especially ChatGPT, have shown exceptional performance in language understanding, reasoning, and interaction, attracting users and researchers from multiple fields and domains. Although LLMs have shown great capacity to perform human-like task accomplishment in natural language and natural image, their potential in handling remote sensing interpretation tasks has not yet been fully explored. Moreover, the lack of automation in remote sensing task planning hinders the accessibility of remote sensing interpretation techniques, especially to non-remote sensing experts from multiple research fields. To this end, we present Remote Sensing ChatGPT, an LLM-powered agent that utilizes ChatGPT to connect various AI-based remote sensing models to solve complicated interpretation tasks. More specifically, given a user request and a remote sensing image, we utilized ChatGPT to understand user requests, perform task planning according to the tasks' functions, execute each subtask iteratively, and generate the final response according to the output of each subtask. Considering that LLM is trained with natural language and is not capable of directly perceiving visual concepts as contained in remote sensing images, we designed visual cues that inject visual information into ChatGPT. With Remote Sensing ChatGPT, users can simply send a remote sensing image with the corresponding request, and get the interpretation results as well as language feedback from Remote Sensing ChatGPT. Experiments and examples show that Remote Sensing ChatGPT can tackle a wide range of remote sensing tasks and can be extended to more tasks with more sophisticated models such as the remote sensing foundation model. The code and demo of Remote Sensing ChatGPT is publicly available at https://github.com/HaonanGuo/Remote-Sensing-ChatGPT .

*Index Terms*— Remote sensing image, large language model, Agent, image interpretation

## 1. INTRODUCTION

The earth observation technique provides an ideal data source for monitoring the ground surface of the earth on a large scale, and thus can support the achievement of the Sustainable Development Goal (SDG)[1]. Over the past decades, tremendous efforts have been made to develop deep learning-based algorithms for a wide range of remote sensing interpretation tasks such as scene classification, object detection, semantic segmentation, image description, etc[2]–[4].

Despite the varied tasks and models that have been developed, how to organize those tasks for solving real-world users' requests remains a challenge[5], [6]. For example, solving the 'Count the number of airplanes on the runway' request needs to perform runway segmentation, airplane detection, and object counting in sequential. This task planning process, however, heavily relies on human intervention currently by asking remote sensing specialists to understand user requests and plan tasks before delivering the product that meets users' expectations. The lack of automation in task planning hinders the accessibility of remote sensing interpretation techniques, especially to non-remote sensing experts from multiple research fields. Recently, large language models (LLM), especially ChatGPT[7], have shown impressive performance in language understanding, reasoning, and interaction. By automatically learning from numerous web text data in an auto-regressive manner, LLM has been proven effective on even unseen tasks[8]. This emergent ability makes LLM possible to perform task planning with a well-designed prompt system. In the natural language and image understanding field, some LLM agent-based methods have proven the feasibility of using LLM as an agent to use tools to process image or language processing tasks[5], [9]. However, the potential of LLM in the remote sensing field has not yet been fully explored. Although some preliminary studies have explored the applicability of ChatGPT in remote sensing tasks [10], they simply apply methods designed for natural images to remote sensing images and have not yet considered integrating remote sensing models with ChatGPT. Moreover, the quantitative evaluation of the task-calling performance of different LLMs has not yet been studied.

In this paper, we propose Remote Sensing ChatGPT, a ChatGPT-like system that is capable of understanding users' requests, planning remote sensing interpretation tasks, and generating final products and responses to users. We built Remote Sensing ChatGPT based on ChatGPT and multiple AI-based remote sensing models that support various interpretation tasks. We expect that Remote Sensing ChatGPT can push forward the accessibility of remote sensing interpretation techniques to non-experts, people who are working on the applications of multiple fields such as urbanization and deforestation but are not remote sensing experts. Furthermore, it is a meaningful attempt to automate remote sensing task planning, one key step toward the realization of fully automatic remote sensing image interpretation. Quantitative and qualitative evaluations are performed to explore the performance of Remote Sensing ChatGPT under different LLM backbones. We also discuss the limitations and future directions of designing a ChatGPT-like system for remote sensing.

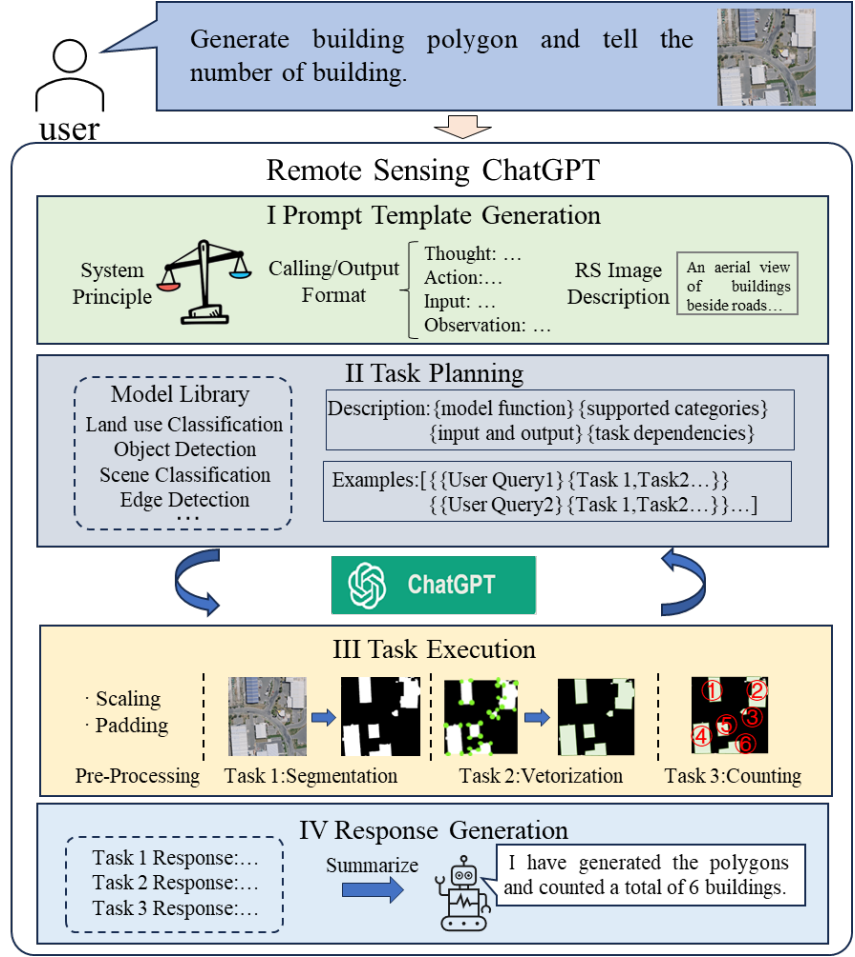

**Fig.1**. Workflow of the proposed Remote Sensing ChatGPT.

## 2. REMOTE SENSING CAHTGPT

Remote Sensing ChatGPT is capable of solving remote sensing tasks with ChatGPT and remote sensing interpretation models. With Remote Sensing ChatGPT, users can simply send a remote sensing image with the corresponding language request, and get the interpretation results and language feedback. The workflow of Remote Sensing ChatGPT is shown in Fig.1, which includes prompt template generation, task planning, task execution, and response generation, which will be introduced in the following subsections.

### 2.1 Prompt Template Generation

Given a user's language input, the first step of Remote Sensing ChatGPT is to generate a prompt template as a system principle for ChatGPT to understand instruct, execute reasoning, and output results correctly. For example, in the system principle part, ChatGPT is required to use tools to finish following tasks, rather than directly imagine from the description. In the task calling and output format part, ChatGPT is told to be strict with the file name and will never fabricate nonexistent files. A template is also provided to ChatGPT to regulate tool names, input files, and output tool observations. Furthermore, considering that ChatGPT is a language model and cannot directly access images, we introduce the BLIP model[11] to caption the remote sensing image and thus provide visual cues for ChatGPT to better understand the image.

### 2.2 Task Planning

Remote Sensing ChatGPT currently supports calling different remote sensing tasks, such as scene classification, land use classification, object detection, image captioning, edge detection, polygonization, and object counting. The details of these tasks and their corresponding models are listed in Table I. We selected the widely used network architecture for each task and trained the models on publicly

Table I Supported Tasks in Remote Sensing ChatGPT

| Tools | Method | Dataset |
|---|---|---|
| Scene Classification | ResNet | AID |
| Land use Classification | HRNet | LoveDA |
| Object Detection | YOLOv5 | DOTA |
| Image Captioning | BLIP | BLIP Dataset |
| Edge Detection | Canny | —— |
| Polygonization | Douglas-Peuker | —— |
| Object Counting | —— | —— |

available benchmarks. It should be noted that more tasks and more sophisticated models can be applied to further improve the performance of Remote Sensing ChatGPT

In this part, given the defined task library, the descriptions of the task functions, supported categories, input and output data format, and task dependencies are generated to supplement the prompt template. Furthermore, some examples of each task are provided to perform in-context learning and thus further improve the model's understanding of user inputs. The complete prompt is then fed into ChatGPT to perform task planning.

**2.3 Task Execution and Response Generation**

The output of ChatGPT determines whether and which tools should be used. The determined tool is applied to the pre-process remote sensing image and generates the corresponding output. The output is then fed to ChatGPT as a new observation to determine whether new tools should be used to further solve the users' requests. If no more tools should be used, the output of all executed tasks will be sent to ChatGPT, which will generate the final response for the user.

## 3. EXPERIMENTS

Considering that Remote Sensing ChatGPT can be easily extended to more tasks with advanced methods such as foundation models, our experiments focus on whether ChatGPT correctly plans the interpretation task rather than the interpretation accuracy. To this end, we collect 138 user queries from multiple users. Then, we label the corresponding tasks based on the queries. Considering that ChatGPT can call more tasks to assist the reasoning process, we label only the essential task for solving user queries and calculate the correctness of the query, representing whether ChatGPT correctly plans the essential task. For example, object detection is the essential task for the query 'locate the baseball diamond in the aerial image provided'.

As shown in Table II, we testified Remote Sensing ChatGPT with 4 different ChatGPT backbones. We find that Remote Sensing ChatGPT with gpt-3.5-turbo achieves the best performance in remote sensing task planning, followed by gpt-4-1106-preview and gpt-4. The overall accuracy of 94.9% demonstrates the capacity of Remote Sensing ChatGPT in understanding user queries and planning remote sensing tasks. Although gpt-3.5-turbo-1106 supports more tokens, its capacity to understand complex instructions is rather limited as compared to gpt-3.5-turbo, leading to degraded model performance.

In Fig.2, we further visualize some successful and failure cases of Remote Sensing ChatGPT. From the successful cases we can see that Remote Sensing ChatGPT can effectively plan and execute not only simple queries that require a single task but also complex queries that require iterative execution of multiple tasks. However, there are also some failure cases as illustrated in Fig.2. One of the main failure cases lies in the unsupported categories of the existing remote sensing models. For example, Remote Sensing ChatGPT requests the land use classification model to

Table II Correctness of Remote Sensing ChatGPT on task planning

| Tools | gpt-4 | gpt-4-1106-preview | gpt-3.5-turbo | gpt-3.5-turbo-1106 |
|---|---|---|---|---|
| **Overall Correctness** | 63% | 84.1% | **94.9%** | 29% |
| Scene Classification | 76.9% | **84.6%** | **84.6%** | 7.7% |
| Land use Classification | 69.1% | 90.9% | **100%** | 30.9% |
| Object Detection | 79.2% | 83.3% | **95.8%** | 33.3% |
| Image Captioning | 86.7% | 60% | **93.3%** | 6.7% |
| Edge Detection | 70% | **100%** | **100%** | **100%** |
| Polygonization | 0% | **100%** | **100%** | 28.6% |
| Object Counting | 0% | 64.3% | **78.6%** | 7.1% |

user.

segment cultivated land from the input image, which however,

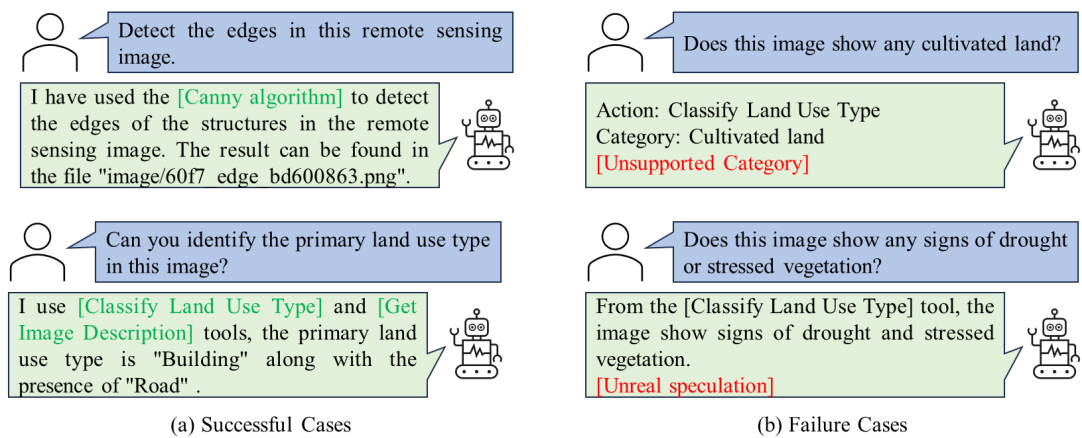

Fig.2 Example of some successful and failure cases of Remote Sensing ChatGPT.

is not supported by the model since the training dataset does not include the cultivated land category. Another failure case shows that Remote Sensing ChatGPT tends to imagine the answer rather than ask for more information when the existing tools or information cannot fully solve the user query.

## 4. FUTURE DIRECTION

Remote Sensing ChatGPT is an LLM-powered agent that utilizes ChatGPT to connect various AI-based remote sensing models to solve complicated interpretation tasks. By coupling remote sensing foundation model and agent-based model, we believe that fully-automated remote sensing interpretation will be achieved in the near future and thus serve users from multiple fields such as environment monitor, disaster response, etc. As Remote Sensing ChatGPT is a preliminary attempt, more research can focus on developing open-vocabulary remote sensing foundation models or parameter-efficient finetuning of LLMs for better performance.

## 5. CONCLUSION

In this paper, we propose Remote Sensing ChatGPT, an LLM-powered agent that utilizes ChatGPT to connect various AI-based remote sensing models and solve complicated interpretation tasks. Remote Sensing ChatGPT can understand users' requests, plan remote sensing interpretation tasks, and generate final products and responses to users. Quantitative and qualitative evaluations have demonstrated that Remote Sensing ChatGPT can perform accurate task planning and execution. We hope that Remote Sensing ChatGPT is a meaningful attempt to realize fully-automated remote sensing image interpretation, and can push forward the accessibility of remote sensing interpretation techniques to researchers who are working on the applications of multiple fields.

## ACKNOWLEDGMENT

This work was supported in part by the National Natural Science Foundation of China under Grants 42230108 and 62371348.